# FINE-TUNING DEEP LEARNING MODELS FOR STEREO MATCHING USING RESULTS FROM SEMI-GLOBAL MATCHING

Hessah Albanwan[1,2] and Rongjun Qin[1,2,3,4,*]

[1] Geospatial Data Analytics Laboratory, The Ohio State University, Columbus, USA
[2] Department of Civil, Environmental and Geodetic Engineering, The Ohio State University, Columbus, USA
[3] Department of Electrical and Computer Engineering, The Ohio State University, Columbus, USA
[4] Translational Data Analytics Institute, The Ohio State University, Columbus, USA
Email: <albanwan.1, qin.324>@osu.edu

Commission VI, WG VI/4

**ABSTRACT:**

Deep learning (DL) methods are widely investigated for stereo image matching tasks due to their reported high accuracies. However, their transferability/generalization capabilities are limited by the instances seen in the training data. With satellite images covering large-scale areas with variances in locations, content, land covers, and spatial patterns, we expect their performances to be impacted. Increasing the number and diversity of training data is always an option, but with the ground-truth disparity being limited in remote sensing due to its high cost, it is almost impossible to obtain the ground-truth for all locations. Knowing that classical stereo matching methods such as Census-based semi-global-matching (SGM) are widely adopted to process different types of stereo data, we therefore, propose a finetuning method that takes advantage of disparity maps derived from SGM on target stereo data. Our proposed method adopts a simple scheme that uses the energy map derived from the SGM algorithm to select high confidence disparity measurements, at the same utilizing the images to limit these selected disparity measurements on texture-rich regions. Our approach aims to investigate the possibility of improving the transferability of current DL methods to unseen target data without having their ground truth as a requirement. To perform a comprehensive study, we select 20 study-sites around the world to cover a variety of complexities and densities. We choose well-established DL methods like geometric and context network (GCNet), pyramid stereo matching network (PSMNet), and LEAStereo for evaluation. Our results indicate an improvement in the transferability of the DL methods across different regions visually and numerically.

## 1. INTRODUCTION

### 1.1 Background

Recently, end-to-end deep learning (DL) networks have been the highlight for stereo matching tasks due to their ability to generate disparity maps directly using a pair of rectified images (Laga, 2019). They have been proven to provide high performance and accuracies due to learning contextual information (Chang & Chen, 2018; Cheng et al., 2020; Kendall et al., 2017). One of the essential criteria to evaluate DL methods in stereo matching is transferability (or generalizability), which refers to the ability to learn and predict from new environments that have not been seen previously by the model in the training dataset. Current studies show that DL stereo matching methods have poor transferability and generalization capabilities if there is a significant domain difference between training and testing datasets (Pang et al., 2018; Song et al., 2021), for example, satellite and ground view sensor images. Typically, we would assume that transferability will enhance if the datasets are from the same source as sensor and have similar characteristics. However, for satellite images, the variation goes beyond color and geometry, the complexity and diversity of urban patterns vary greatly across cities around the world, where each region has its unique spatial distribution of buildings, landscapes, roads, etc., in addition, to the buildings heights, shapes, etc. We strongly believe that these spatial variations limit the usage of DL methods in remote sensing tasks and may lead to poor performance and worse transferability.

Typically, DL methods require a large amount of training data to enhance the transferability and provide good accuracies. For remote sensing data, even with satellite images becoming more accessible, the ground truth data (as Light detection and ranging (LiDAR)) remain scarce. Some of the existing benchmarks provide the ground truth LiDAR for several regions around the world, but they are unevenly distributed around the world covering mostly major cities. With ground truth data being limited, current studies struggle to find alternative solutions for this problem, leaving this question open to be addressed.

As opposed to DL stereo matching algorithms, Census cost metric has been proved to provide a good trade-off between the accuracy and robustness to extremely varying scenes and sensors (Chen et al., 2019; Loghman & Kim, 2013). However, it may not produce optimal results as DL methods when they are sufficiently trained and tested on similar datasets. Additionally, unlike DL stereo matching methods, Census requires cost aggregation using methods like semi-global matching (SGM) to further enhance the results. Given these properties of Census-SGM, we believe that it can be a good candidate to replace the ground-truth disparity map during the learning process. However, Census-SGM is known to perform poorly in some regions (e.g., texture-less regions) (Humenberger et al., 2010), therefore, masking reliable pixels using complementary information like SGM's energy map and the texture or edge map is important to be able to use Census-SGM as ground truth. Therefore, we propose a method that can learn directly from the target dataset using the disparity map from Census-SGM to only compare pixels with low uncertainty.

In this paper, we aim to tackle two major challenges of DL methods used in stereo matching and remote sensing satellite images, the transferability and lack of ground truth data. We first, analyse the transferability of the DL methods across 20 study sites around the world by considering three of the most used well-established methods including geometric and context network (GCNet), pyramid stereo matching (PSMNet), and LEAStereo. We then use the pre-trained satellite models for finetuning, where we only modify the loss function to compare between the predicted and Census-SGM disparities around pixels with high confidence (i.e., low uncertainty). We evaluate our results qualitatively and quantitatively based on the triangulated digital surface models (DSM). In summary, our paper provides the following contributions:

- We analyse the transferability and performance of the DL methods used for stereo matching across satellite images from different regions.
- We provide a fine-tuning solution using census-SGM to improve the transferability/generalizability of DL methods by learning directly from the target dataset without requiring ground truth data.

---

[*] Corresponding author

The remainder of the paper is organized as follows, Section 1.2 discusses the related works and rationale of this paper, Section 2, provides an overview of the datasets and stereo methods used in this work, Section 3, the analysis, discussion, and results, and finally, we provide the conclusion and potential future works in Section 4.

**1.2 Related Works**

**1.2.1 Deep learning methods for stereo matching**
State-of-the-art DL methods have been reported to have remarkable progress in stereo matching. Unlike classical methods where disparity maps have to be generated on four different stages (i.e., cost matching, aggregation, disparity computation, and refinement) (Scharstein et al., 2001), deep learning networks have provided an alternative for a direct solution that integrates all steps into a single network to predict the disparity map using a pair of stereo images. (Mayer et al., 2016) are the first to propose an deep learning network for stereo matching, their method consists of 2D convolutional neural networks (CNNs) embedded in an encoder-decoder Siamese network; they applied their method on KITTI dataset and proved to be able to get high accuracy than Census-SGM and DL networks as MC-CNN (Žbontar & LeCun, 2015). Later on, geometry was realized as an added value to the stereo matching problems, Kendall et al., (2017) in their geometry and context network (GCNet) have found that learning from geometry can greatly enhance the disparity map. The development of DL methods is still in progress and many current algorithms are inspired by previous advances in stereo matching (Chang & Chen, 2018; Cheng et al., 2020), however, due to their novelty, there still exist some limitations that need to be addressed. One of the major challenges for DL methods is the limited transferability and generalization capabilities (Pang et al., 2018; Song et al., 2021). As supervised methods, they cannot predict beyond the training data leading to poor performance around unseen instances. Additionally, they are often computationally expensive demanding a substantial amount of training time and data to assure high performance (Knöbelreiter et al., 2019) and good transferability. However, a large amount of training data is time-consuming and often unavailable for remote sensing data.

**1.2.2 The challenges of remote sensing data used in stereo image matching**
The outstanding performance of DL methods on standard benchmarks such as KITTI, Middlebury, and Sceneflow have inspired many other fields like remote sensing. Some studies have shown that they can have high accuracy than classic stereo matching algorithms when applied to satellite images (M. Chen et al., 2021), however, their performance is limited by the number of training datasets and computational time. Compared with ground-view, synthetic, or camera images, remote sensing images are more complex, and even with the continuous development of high-resolution sensors, remote sensing images are rich with information and dense in content leading to a more difficult stereo matching. The complexity in remote sensing images come from various aspects including, 1) varying image content based on numerous regions around the world, this leads to highly varying spatial patterns, landscapes, textures, etc., 2) characteristics of objects in the image, for example, size, height, and shape of buildings (Qin, 2014), and 3) existence of occlusions from shadows or view angles (Qin, 2019). These variabilities in remote sensing images may highly compromise the performance and transferability of DL methods.

Another major challenge in remote sensing is the lack of ground truth data (Cournet et al., 2020). Typically, the ground truth disparity map for training the networks for stereo matching is often transformed and generated from Light Detection and Ranging sensor (LiDAR), however, due to its expensive price, it is impossible to collect LiDAR data for all locations around the globe. With this being an issue, this limits the capabilities of deep learning networks in many remote sensing tasks.

**1.2.3 Transferability solutions for DL networks**
It is a general practice that a model may be re-trained or fine-tuned when applied to unseen data. In the absence of the ground truth data re-training from scratch may not be a viable option. Instead of having ground truth data for training, studies have indicated that finetuning can be performed on alternatives to the ground truth data, for example, (Zhong et al., 2017) proposed a self-supervised DL method that used image warping errors for the loss function in the training stage. Similarly, (Zhou et al., 2017) proposed an unsupervised learning method that takes the loss from the left-right consistency check. Many other studies suggest unsupervised or self-supervised learning algorithms for finetuning the pretrained model without requiring the ground truth data (Knöbelreiter et al., 2019; Yuan et al., 2021). Hence, in this work, we propose a training approach that allows our network to adapt to new datasets without demanding the ground truth. This can be applied by training the target domain with Census-SGM as an alternative for the ground truth. Census cost metric can work well in most environments and SGM can provide the energy map to indicate reliable pixels with low uncertainty.

**2. METHODOLOGY**

In this section, we describe the data used in the experiment, present the relevant background information about the current stereo matching DL methods, and discuss the proposed finetuning method.

**2.1 Dataset Description**
*Source dataset* is the training dataset that has available ground-truth disparity maps. We use this dataset for the pre-trained model. The source data is from the 2019 Institute of Electrical and Electronics Engineers (IEEE) Geoscience and Remote Sensing Society (GRSS) Data Fusion Contest (**DFC**) track 2 benchmark (Le Saux, 2019); it includes 4320 rectified satellite images from Worldview-3 sensor with their ground truth disparity maps and dimensions of 1024 x 1024. We pre-process by normalizing and cropping the images to smaller patches of size 1248 x 384 to reduce memory consumption, thus, in total, we have 25,000 training patches, we divide the patches into 80% for training and 20% for testing.

***Target dataset*** is used for the proposed finetuning method. It is composed of stereo pair satellite images from different sensors and regions around the world. We select 20 study sites from around the world to cover a wide range of land covers, buildings types, shapes, heights (e.g., residential areas, industrial areas, etc.), densities, and complexities. Our study sites cover few sub-areas in 1) Omaha, Nebraska, USA, 2) Jacksonville, Florida, USA, 3) near San Fernando Argentina, 4) London, England, 5) Haiti, and 6) Rochor and Punggol, Singapore. The stereo pair images are very high resolution (VHR) satellite images from different sensors including Worldview-3, Worldview-2, IKONOS, and GeoEye-1. We pre-process the stereo pair images and unify the spatial resolution by up-sampling to 0.3 meters, we then rectify each pair of images to be processed by the stereo matching algorithms. For more information about the datasets refer to Figure 1.

For finetuning in the target domain, we use the stereo pairs from target datasets with their Census-SGM disparity and energy maps. We finetune using 571 training and 393 testing patches of size (1248 x 384). Since we are using the pre-trained model from the source data for finetuning we only need a small sample from the target data to finetune.

Some benchmarks provide the ground-truth DSM which we use for evaluation of the proposed finetuning methods. The study sites with the ground truth DSM include Omaha and Jacksonville datasets from the 2019 DFC benchmark, Argentina dataset provided by the **IARPA** (The Intelligence Advanced Research Projects Activity) Multiview stereo 3D mapping challenge, and London dataset.

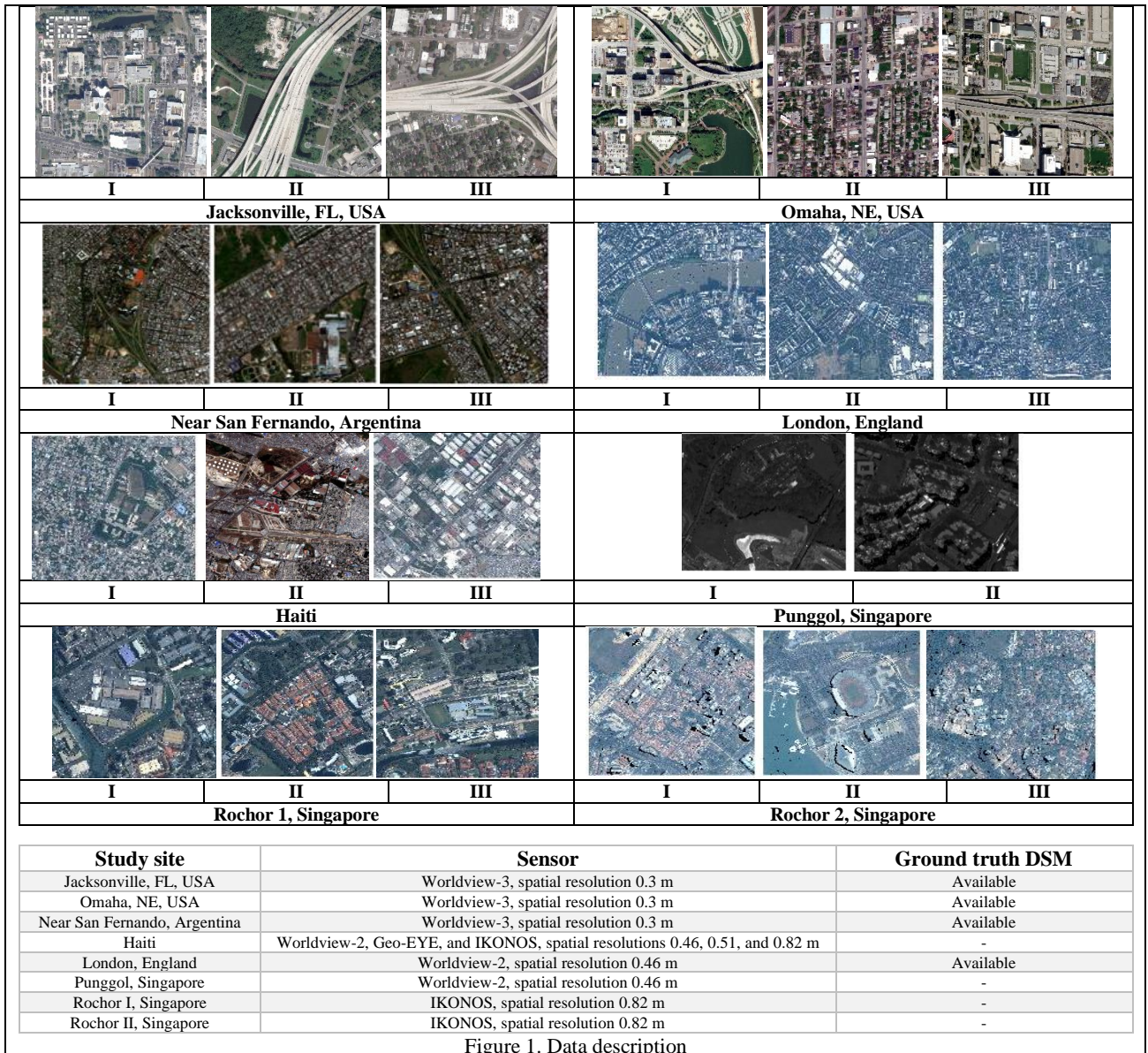

| Study site | Sensor | Ground truth DSM |
|---|---|---|
| Jacksonville, FL, USA | Worldview-3, spatial resolution 0.3 m | Available |
| Omaha, NE, USA | Worldview-3, spatial resolution 0.3 m | Available |
| Near San Fernando, Argentina | Worldview-3, spatial resolution 0.3 m | Available |
| Haiti | Worldview-2, Geo-EYE, and IKONOS, spatial resolutions 0.46, 0.51, and 0.82 m | - |
| London, England | Worldview-2, spatial resolution 0.46 m | Available |
| Punggol, Singapore | Worldview-2, spatial resolution 0.46 m | - |
| Rochor I, Singapore | IKONOS, spatial resolution 0.82 m | - |
| Rochor II, Singapore | IKONOS, spatial resolution 0.82 m | - |

Figure 1. Data description

**2.2 The stereo matching deep learning methods**
DL methods provide direct solutions to stereo matching problems, where it takes a pair of rectified images and outputs the disparity map without further optimization as cost aggregation. We use three common DL methods that have been reported to have an outstanding performance in stereo matching computer vision tasks including CGNet, PSMNet, and LEAStereo. In this section, we briefly explain the mechanism of these methods.

*A. Geometry and context network (GCNet)*
GCNet is developed by Kendall et al. (2017); they develop a Siamese network for their 2D convolution with shared weights to extract unary features from pair of images and generate the 4D cost volume (height, width, disparity, and feature size). The high dimension cost volume enables geometry to be preserved during cost matching. Additionally, instead of using subtraction or distance metrics to compute the cost, they rely on feature concatenation from both images, allowing better performance. They also implement a regularization step using 3D convolution to produce better feature representation using image dimensions and the disparity as an additional dimension.

**B.  Pyramid stereo matching (PSMNet)**

PSMNet learns the cost volume by integrating contextual feature cues from different scales; they construct spatial pyramid pooling (SPP) module to expand learning from pixel-level features to regional-level features on various scales. They build a stack hour-glass module to learn and regularize the 4D cost volume using 3D convolutional layers.

**C.  LEAStereo**

As an advancement to previous methods, LEAStereo build an automated neural architectural search (NAS) module to find optimal parameters (e.g., filter size, strides, etc.) for the convolutional layers in the networks. Their stereo matching algorithm consists of a feature extraction network and cost matching network.

### 2.3 Proposed finetuning method

**A.  Semi-global matching (SGM)**

SGM is originally proposed to impose a smoothness constraint on the matched cost. Its energy function can be expressed as follows:

$$E(D) = \sum_{x}(C(x,d) + \sum_{q \in N_p} P_1\, T[|d_x - d_y| = 1] + \sum_{q \in N_p} P_2\, T[|d_x - d_y| > 1]) \qquad (1)$$

Where the $C(x,d)$ is the computed cost for the pixel at position x and disparity d. The $P_1$ and $P_2$ are the penalty parameters defined by users and are applied to regularize the cost function for all pixels $q$ in the neighbourhood $N_p$. $P_1$ is a constant value for pixels, it accounts for pixels on slanted regions, whereas $P_2$ is proposed to impose large values on pixels discontinuities.

The key idea for SGM (Hirschmuller, 2005, 2008) is to aggregate the matching cost iteratively from multiple directions. The cost aggregation can be applied in eight or sixteen directions depending on the desire of users. The eight directions can provide fair results and fast process, while the sixteen directions can produce more accurate results but at the expense of speed. The accumulated cost from SGM into a single pixel can be expressed as follows:

$$S(x,d) = \sum_r L_r(x,d) \qquad (2)$$

Where $S(x,d)$ is the total accumulated cost from all costs $L_r(x,d)$ at different directions $r$ for pixel at position x and disparity d. Each cost is computed as follows:

$$\begin{aligned}L_r(x,d) = C(x,d) + \min\{&L_r(x-r,d),\ L_r(x-r,d-1) + P_1,\ L_r(x-r,d+1) \\ &+ P_1,\ \min_i L_r(x-r,i) + P_2\} \\ &- \min_k L_r(x-r,k)\end{aligned} \qquad (3)$$

Where the $C(x,d)$ is the computed cost for the pixel x, $P_1$ and $P_2$ are the penalty parameters that are user-defined and are applied to regularize the cost function. We refer to the minimal aggregated/accumulated cost $S(x,d)$ as the energy map in which we use to guide the training of the DL algorithms. A sample for the energy map is shown in Figure 2.

**B.  Proposed finetuning with SGM**

We suggest a generic finetuning approach that can be applied to any DL stereo matching algorithm. We introduce a method that can learn from the target data without having their ground truth disparity maps for training. Instead, we use Census-SGM as the ground-truth disparity. Census-SGM provides a good trade-off for robustness and accuracy, as it has stable performance across datasets of different sensors and regions. However, as prediction method, it does not work well in some areas like flat and texture-less regions. Therefore, we propose a weighted loss function that assigns weights based on the importance of the information. We extract confidence maps which we regard as the most important features we want to learn from. We have two confidence maps, 1) the edge map which we extract from the left image using Canny edge detector to get the edges and textured regions, and 2) the energy map from Census-SGM to indicate pixels with low uncertainty. For information about the confidence maps refer to Figure 2.

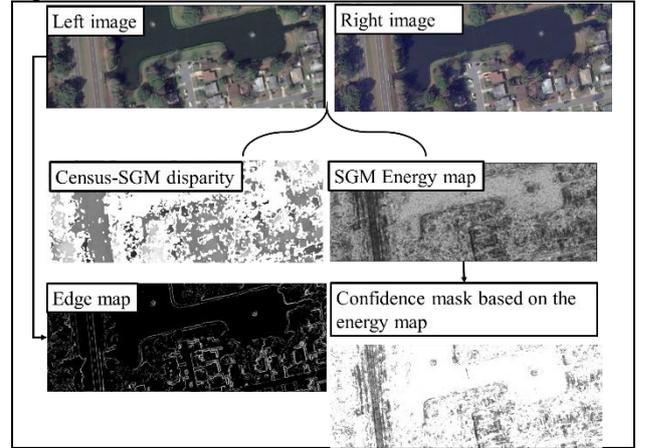

Figure 2. Confidence maps extracted from Census-SGM's energy map and the edge map from left image

In addition to training using the target dataset, we have also applied some data augmentation on some of the source datasets for regularization and smoothness for the results. The architecture of the networks remains intact, the only modification we apply is on the loss function. The modified loss function is expressed as follows:

$$Loss = Loss_{augumneted\ source} + Loss_{target} \qquad (4)$$

Where the $Loss_{augumented\_source}$ computes the L1 loss function between the augmented source dataset and their provided ground truth disparity. The $Loss_{target}$ computes the loss for the target data by comparing the predicted and Census-SGM disparities. We use Huber loss for the target loss because of its robustness to outliers. To optimize the loss function, we divide our target loss into three parts: first, the loss for all pixels in the predicted disparity against the Census-SGM disparity, which can be expressed as follows:

$$Loss_1 = huber\_loss \begin{bmatrix} Census\ disparity, \\ predicted\ disparity \end{bmatrix} \qquad (5)$$

Second, the loss for pixels with high confidence based on the energy map. We apply a threshold on the energy map to mask the pixels with low uncertainty. We then use Huber loss function as follow:

$$Loss_2 = huber\_loss \begin{bmatrix} Census\ disparity(mask_{energyemap}), \\ predicted\ disparity\ (mask_{energyemap}) \end{bmatrix} \qquad (6)$$

Where $mask_{energymap} = ENERGY\_MAP < threshold$. The threshold is chosen empirically and set to 2500. Third, the loss for pixels with high confidence based on the texture and edge map which we extract using canny edge detection. The Huber loss is then computed based on the valid pixels masked from the edge map as follows:

$$Loss_3 = huber\_loss \begin{bmatrix} Census\ disparity(mask_{edgemap}), \\ predicted\ disparity\ (mask_{edgemap}) \end{bmatrix} \quad (7)$$

We combine these three losses into the target loss. Since Census-SGM do not work well for all areas (e.g., flat, texture-less, etc.), we further optimize the loss function by imposing weights based on the importance of information and likelihood that it will have high confidence. We assign low weights for $Loss_1$ and high weights for $Loss_2$ and $Loss_3$ weights. The final target function can be expressed as follows:

$$Loss_{target} = w_1 * Loss_1 + w_2 * loss_2 + w_3 * loss_3 \quad (5)$$

The weights are determined empirically and set to $w_1 = 0.1, w_2 = 0.45, w_3 = 0.45$.

## 3. EXPERIMENT AND RESULTS

To evaluate the transferability of DL stereo matching algorithms using satellite images, we triangulate the disparity maps to DSMs. First, we visually analyse the transferability of DL algorithms across different study sites before finetuning. Then, we present visual results and numerical evaluation for before and after finetuning by computing the root mean square error (RMSE) against the ground truth DSM.

### 3.1 DL training setup

GCNet and PSMNet are implemented in Pytorch, they use Adam optimizer with β1=0.9 and β2=0.999, the initial learning rate of 0.001, and training and testing batch size of 1. For training, the images are randomly cropped to 256 x 512. As for LEAStereo, it is also implemented in Pytorch, the training images cropped into 288 x 576, we use SGD optimizer with β1=0.9, a cosine learning rate ranging from 0.025 to 0.001, and a batch size of 1.

### 3.2 Visual analysis before fine-tuning

The visual analysis allows viewing and inspecting the results based on the visible characteristics to make informative decisions and determinations about behavioural changes, consistency, variability, and overall accuracy of the data. In this section, we present the DSMs generated by GCNet, PSMNet, and LEAStereo before finetuning across different study sites, we expect their performance to vary based on region.

We present a visual comparison for the DSMs generated by Census-SGM, GCNet, PSMNet, and LEAStereo from different areas in the world in Figure 3. Our results show that DL methods work well for areas with sparse spatially distributed objects that have large buildings, as can be seen from Omaha, Jacksonville, and Rochor, Singapore study sites in Figure 3, where we can see clear outlines for different classes. While for dense areas with many buildings as in London, Argentina, and Haiti study sites, we can see that the predictions are not as meaningful and distinct as in the other study sites. For example, in London study site, we can barely distinguish between different classes in the DSMs of the three DL algorithms (see last raw Figure 3). In some cases as in the Argentina study site, we can see that the DSM is very smoothed especially around dense residential areas, this can be apparent in PSMNet and LEAStereo, while for the Haiti dataset with even more dense areas having small-sized residential houses, the prediction of DL algorithms is worst. In contrast, we can see that Census-SGM has stable performance across all presented study sites, and even with dense areas, we can still distinguish between different classes as in London and Haiti study sites we can clearly differentiate between roads, ground, and buildings, while in the DL algorithms they are barely recognizable. Therefore, we can conclude that the performance for DL stereo matching algorithms is highly sensitive to the variances in sensors and regions.

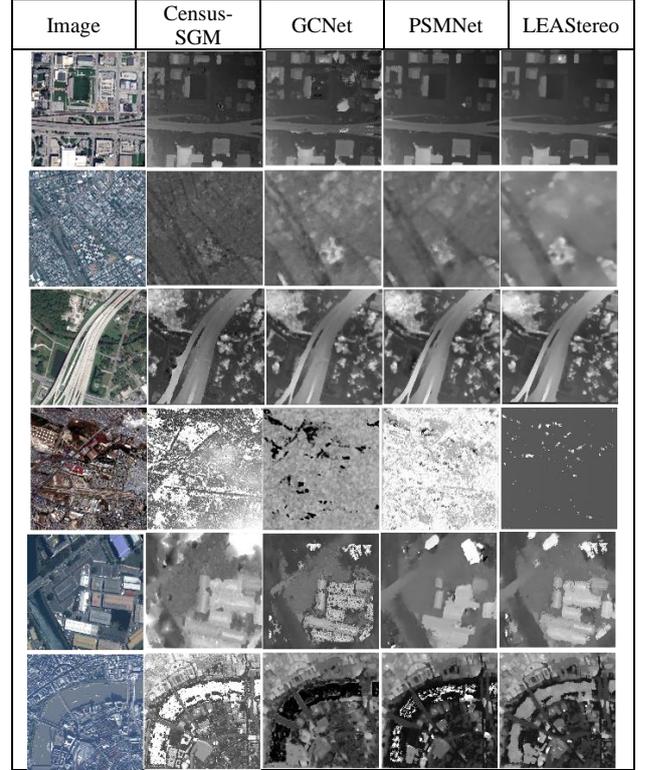

Figure 3. Visual analysis for the transferability of the DL algorithms across different study sites. Note: the rows from 1-6 show Omaha I, Argentina III, Jacksonville II, Haiti II, Rochor I Singapore, and London study sites, respectively.

### 3.3 Finetuning results

We evaluate the DSMs generated by the GCNet, PSMNet, and LEAStereo before and after the proposed finetuning. We use analyze the results visually and numerically by evaluating the DSMs and their RMSE for the study sites that have the ground-truth DSM available including Omaha, Jacksonville, Argentina, and London. We anticipate that the level of details will enhance as we add more information from Census-SGM energy map and disparity map.

We present sample DSMs for the before and after finetuning the DL algorithms for London and Haiti study sites in Figure 4. and Figure 5, respectively. We can notice that the level of details in London study site enhanced significantly which is visible from the enlarged highlighted region in red. Before finetuning the roads and small buildings are very blurry and almost impossible to distinguish, whilst after finetuning with Census-SGM, we can see that the different classes outlined clearly like the ground, roads, and small buildings. For Haiti study site, before finetuning the DSMs are inaccurate and invisible (see Figure 5), however, after finetuning, we can see the details of the city more obvious in all DSMs from GCNet, PSMNet, and LEAStereo (see red rectangle in Figure 5). This indicates that finetuning with Census-SGM and target data can indeed enhance the results.

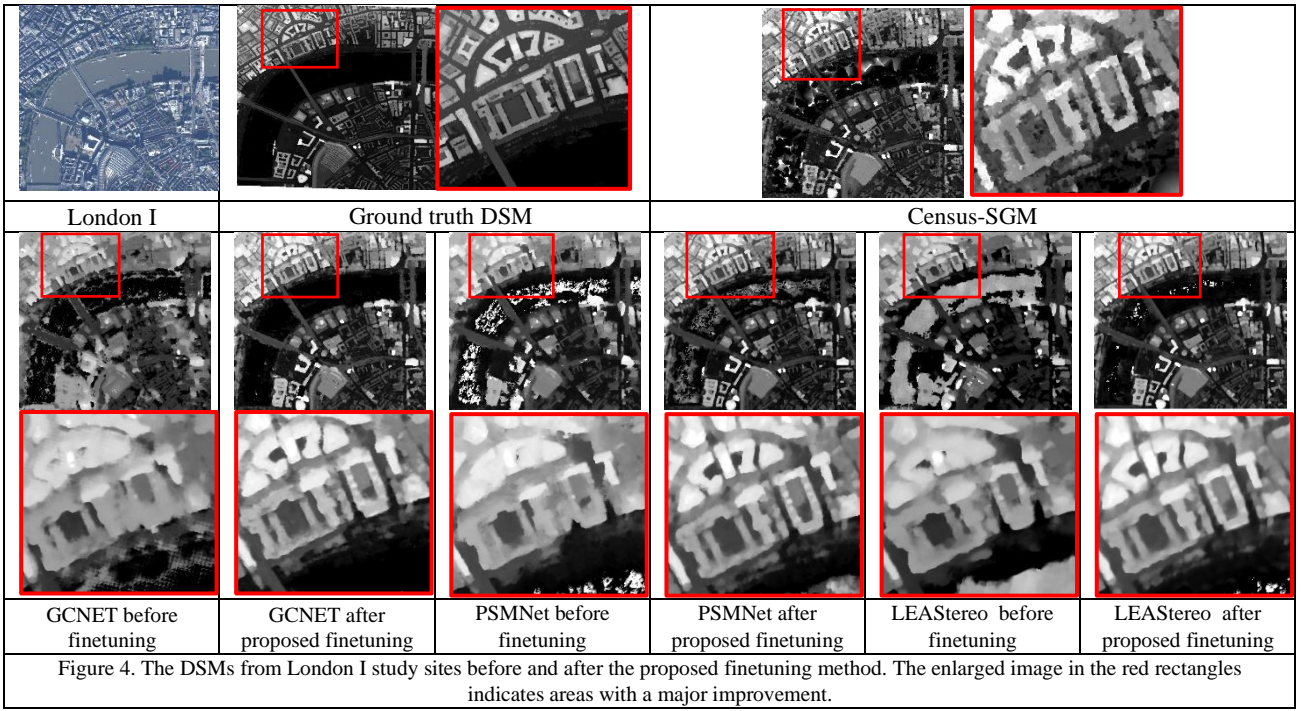

Figure 4. The DSMs from London I study sites before and after the proposed finetuning method. The enlarged image in the red rectangles indicates areas with a major improvement.

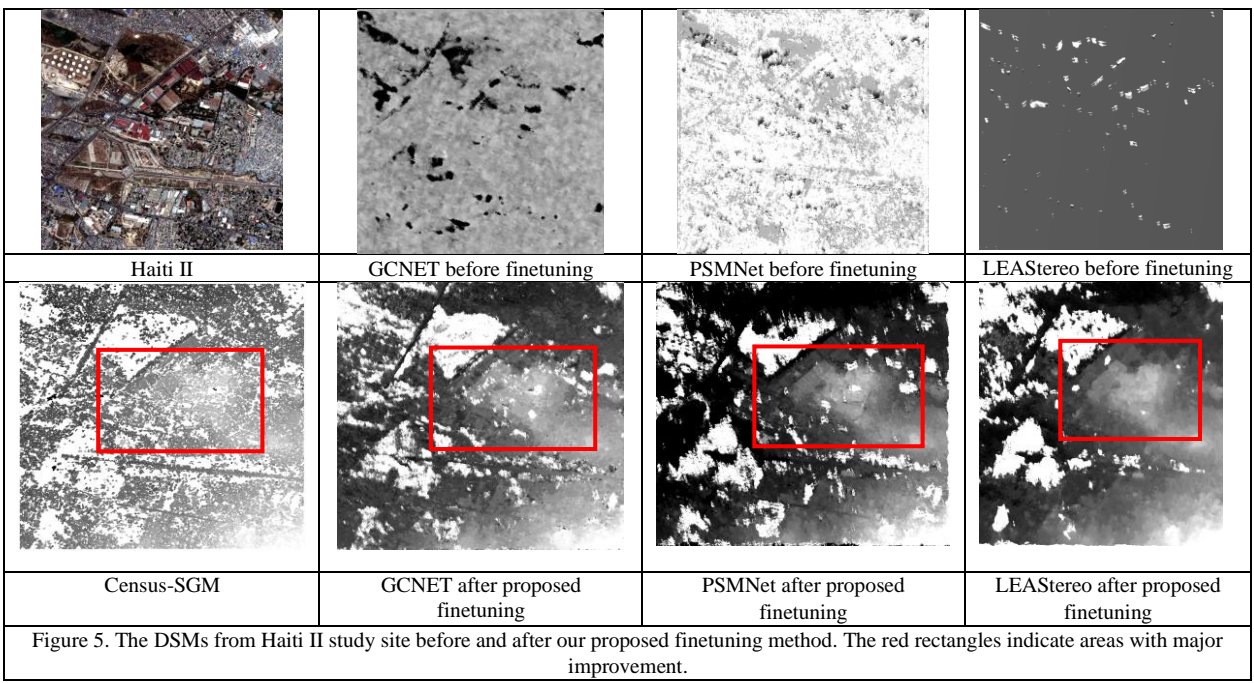

Figure 5. The DSMs from Haiti II study site before and after our proposed finetuning method. The red rectangles indicate areas with major improvement.

We present the numerical evaluation of the RMSE for the DSMs generated by GCNet, PSMNet, and LEAStereo before and after the proposed finetuning approach in two parts: first, the profile that shows the alignment between the ground-truth DSM and the DL algorithms before and after finetuning, second, Table I to show the error (i.e., RMSE) for the DSM in each study sites with the overall average RMSE.

Figure 6. shows the DSM profile for a line segment from London I study site comparing between the ground-truth DSM and the DSMs from GCNet, PSMNet, and LEAStereo before and after finetuning. From the profile of GCNet and PSMNet, we can notice that the DSMs after finetuning become close to the shape and pattern of the ground truth DSM (See red and green lines), whereas for before finetuning the blue line is almost straight and does not reflect much of the true values or patterns form the ground-truth DSM. LEAStereo profile looks much better than before finetuning, and it shows more matching pattern to the ground-truth DSM than GCNet and PSMNet, this is clearly obvious in the middle section from curves showing the drops between the edges and boundaries and flat regions. We can see clear separation of DSM values.

Table I presents the errors (RMSE) for the DSMs from GCNet, PSMNet, and LEAStereo before and after finetuning against the Census-SGM. We evaluate the study sites that have the ground-truth DSM available like Argentina, Omaha, Jacksonville, and London. In general, we can see a notable improvement in the results after the proposed finetuning method, this can be observed from few aspects. First, the RMSE for the DSMs finetuned are lower than before finetuning. This can be seen by the difference (Δ) column in Table I that shows the RMSE of the after finetuning minus raw results (before finetuning); the negative values indicate a drop in the RMSE. The overall average drop in the RMSE for GCNet, PSMNet, and LEAStereo range between 0.07 and 5.3 meters. Second, we can see that that the minimum RMSE for all study sites always falls in one of the finetuned DL methods (See bold numbers in Table I). Third, the average RMSE shows that the lowest errors are achieved by the finetuned DL algorithms of less than 4.80 meters, which are also less than Cesus-SGM whose average is 5.33 meters. In general, the RMSE values of the finetuned DSMs are better than Census-SGM, where they are lower by 0.02 to 3.29 meters.

These results imply that proposed finetuning method using target data and Census-SGM can improve the performance of stereo matching numerically and visually.

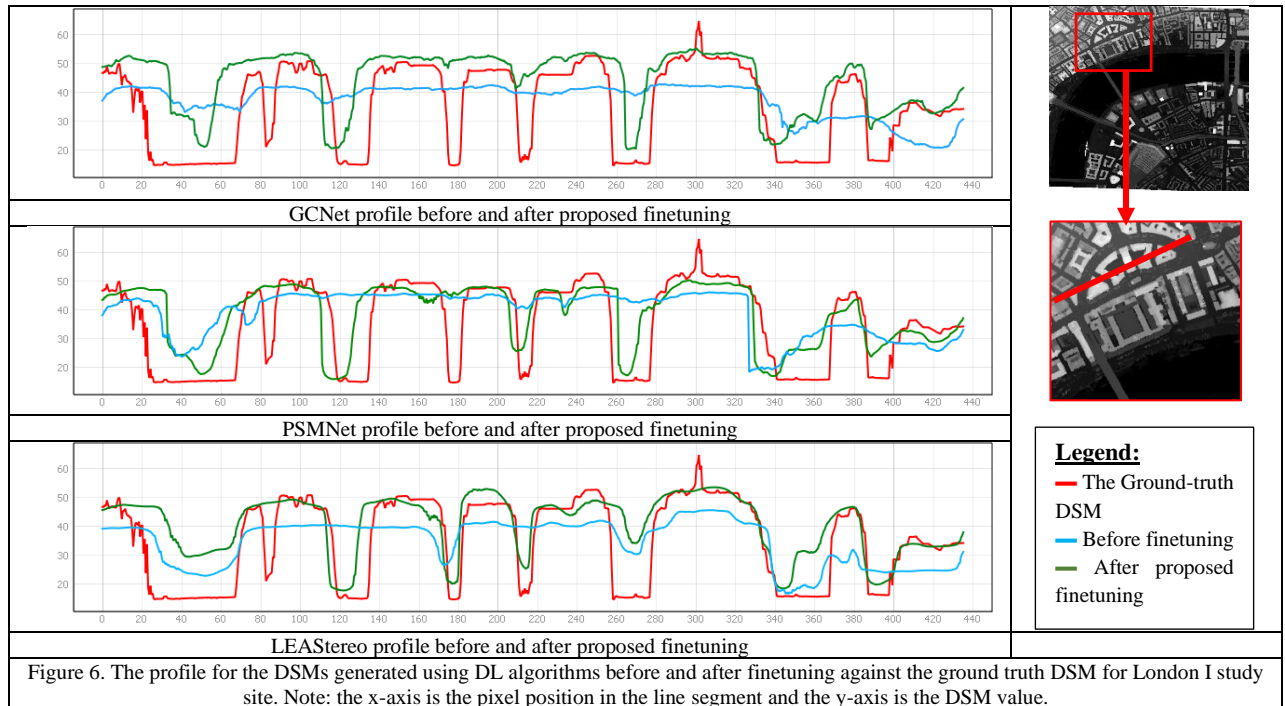

Figure 6. The profile for the DSMs generated using DL algorithms before and after finetuning against the ground truth DSM for London I study site. Note: the x-axis is the pixel position in the line segment and the y-axis is the DSM value.

Table I. Numerical evaluation for the RMSE (meters) of DSMs generated using different DL algorithms for all study sites. Note: the Δ refers to the difference between the RMSE for DL algorithms before and after finetuning.

| Study site | Census-SGM | GCNet | GCNet-finetuned | Δ [GCNet finetuned-GCNet] | PSMNet | PSMNet finetuned | Δ [PSMNet finetuned-PSMNet] | LEAStereo | LEAStereo finetuned | Δ [LEAStereo finetuned-LEAStereo] |
|---|---|---|---|---|---|---|---|---|---|---|
| Argentina I | 5.86 | 5.40 | 5.22 | -0.18 | 5.17 | 4.99 | -0.18 | 4.70 | **4.37** | -0.33 |
| Argentina II | 6.53 | 3.87 | 3.44 | -0.43 | 4.99 | 4.46 | -0.53 | 4.57 | **3.24** | -1.33 |
| Argentina III | 4.23 | 3.78 | **3.51** | -0.27 | 4.02 | 3.66 | -0.36 | 3.64 | 3.53 | -0.11 |
| Omaha I | 3.86 | 4.11 | 3.60 | -0.51 | 2.63 | **3.29** | 0.66 | 3.05 | 3.84 | 0.79 |
| Omaha II | 5.16 | 6.69 | 6.74 | 0.05 | 4.90 | **4.15** | -0.75 | 4.49 | 4.75 | 0.26 |
| Omaha III | 3.23 | 3.18 | 3.94 | 0.76 | 2.84 | **2.24** | -0.60 | 2.69 | 2.80 | 0.11 |
| Jacksonville I | 5.02 | 5.03 | 4.84 | -0.19 | 4.34 | **4.95** | 0.61 | 5.46 | 6.06 | 0.60 |
| Jacksonville II | 4.38 | 3.38 | 3.16 | -0.22 | 3.08 | **2.91** | -0.17 | 3.19 | 3.60 | 0.41 |
| Jacksonville III | 3.79 | 2.82 | **2.60** | -0.22 | 3.00 | 2.93 | -0.07 | 2.68 | 3.15 | 0.47 |
| London I | 9.03 | 13.77 | 9.74 | -4.03 | 14.13 | **8.82** | -5.30 | 9.61 | 8.91 | -0.70 |
| London II | 6.93 | 8.48 | 6.34 | -2.14 | 5.91 | 5.67 | -0.25 | 6.14 | **5.28** | -0.87 |
| London III | 5.93 | 5.74 | 4.75 | -0.99 | 5.02 | **4.67** | -0.36 | 5.06 | 4.78 | -0.28 |
| Average RMSE (m) | 5.33 | 5.52 | 4.82 | -0.70 | 5.00 | **4.40** | -0.61 | 4.61 | 4.53 | -0.08 |

## 4. CONCLUSION AND FUTURE WORKS

In this work, we have examined the possibility of integrating the high-performance DL stereo matching methods with the stable Census-SGM as an unsupervised learning approach to enhance transferability across different study sites. We have analysed the performance of DL methods across different regions and found that they are highly sensitive to unseen regions and environments, which influence the quality of the results. We have proposed a finetuning approach that focus on solving the lack of ground-truth data for satellite images used for stereo matching. It therefore aims to adapt and train using information directly from the target data and Census-SGM. The finetuning has improved the results visually and numerically. The overall average drop in the RMSE is from 0.07 to 5.3 meters after finetuning. While the drop between the finetuned and Census-SGM results is between 0.02 to 3 meters. Our approach shows high potential and needs to be thoroughly investigated in the future to take advantage of the complementary benefits of both methods.

## 5. ACKNOWLEDGEMENTS

The study is partially supported by the ONR grant (award no. N000142012141). The author would like to thank the providers of the benchmark dataset including John Hopkins university applied physics lab to support the imagery, the IARPA to organize the 3D challenge and the IEEE GRSS Image Analysis and Data Fusion Technical Committee.

## 6. REFERENCES

Chang, J.-R., & Chen, Y.-S. (2018). Pyramid Stereo Matching Network. *Proceedings of the IEEE Conference on Computer Vision and Pattern Recognition*, 5410–5418.

Chen, B., Qin, R., Huang, X., Song, S., & Lu, X. (2019). *A Comparison of Stereo-Matching Cost Between Convolutional Neural Network and Census for Satellite Images*. 7. rXiv preprint arXiv:1905.09147.

Chen, M., Briffa, J. A., Valentino, G., & Farrugia, R. A. (2021). Stereo matching of remote sensing images using deep stereo matching. *Image and Signal Processing for Remote Sensing XXVII*, 11862, 60–65. https://doi.org/10.1117/12.2597702

Cheng, X., Zhong, Y., Harandi, M., Dai, Y., Chang, X., Drummond, T., Li, H., & Ge, Z. (2020). Hierarchical Neural Architecture Search for Deep Stereo Matching. *Advances in Neural Information Processing Systems*, 33.

Cournet, M., Sarrazin, E., Dumas, L., Michel, J., Guinet, J., Youssefi, D., Defonte, V., & Fardet, Q. (2020). Ground Truth Generation and Disparity Estimation for Optical Satellite Imagery. *The International Archives of the Photogrammetry, Remote Sensing and Spatial Information Sciences*, XLIII-B2-2020, 127–134. https://doi.org/10.5194/isprs-archives-XLIII-B2-2020-127-2020

Hirschmuller, H. (2005). Accurate and Efficient Stereo Processing by Semi-Global Matching and Mutual Information. *2005 IEEE Computer Society Conference on Computer Vision and Pattern Recognition (CVPR'05)*, *2*, 807–814 vol. 2.

Hirschmuller, H. (2008). Stereo Processing by Semiglobal Matching and Mutual Information. *IEEE Transactions on Pattern Analysis and Machine Intelligence*, *30*(2), 328–341. https://doi.org/10.1109/TPAMI.2007.1166

Humenberger, M., Engelke, T., & Kubinger, W. (2010). A census-based stereo vision algorithm using modified Semi-Global Matching and plane fitting to improve matching quality. *2010 IEEE Computer Society Conference on Computer Vision and Pattern Recognition - Workshops*, 77–84. https://doi.org/10.1109/CVPRW.2010.5543769

Kendall, A., Martirosyan, H., Dasgupta, S., Henry, P., Kennedy, R., Bachrach, A., & Bry, A. (2017). End-to-End Learning of Geometry and Context for Deep Stereo Regression. *2017 IEEE International Conference on Computer Vision (ICCV)*, 66–75. https://doi.org/10.1109/ICCV.2017.17

Knöbelreiter, P., Vogel, C., & Pock, T. (2019). Self-Supervised Learning for Stereo Reconstruction on Aerial Images. *ArXiv:1907.12446 [Cs]*. http://arxiv.org/abs/1907.12446

Laga, H. (2019). A Survey on Deep Learning Architectures for Image-Based Depth Reconstruction. *ArXiv:1906.06113 [Cs, Eess]*.

Le Saux, B. (2019). *Data Fusion Contest 2019 (DFC2019)* [Data set]. IEEE DataPort. https://doi.org/10.21227/C6TM-VW12

Loghman, M., & Kim, J. (2013). Sgm-Based Dense Disparity Estimation Using Adaptive Census Transform. *2013 International Conference on Connected Vehicles and Expo (ICCVE)*, 592–597. https://doi.org/10.1109/ICCVE.2013.6799860

Mayer, N., Ilg, E., Hausser, P., Fischer, P., Cremers, D., Dosovitskiy, A., & Brox, T. (2016). A large dataset to train convolutional networks for disparity, optical flow, and scene flow estimation. *2016 IEEE Conference on Computer Vision and Pattern Recognition (CVPR)*, 4040–4048. https://doi.org/10.1109/CVPR.2016.438

Pang, J., Sun, W., Yang, C., Ren, J., Xiao, R., Zeng, J., & Lin, L. (2018). Zoom and Learn: Generalizing Deep Stereo Matching to Novel Domains. *2018 IEEE/CVF Conference on Computer Vision and Pattern Recognition*, 2070–2079. https://doi.org/10.1109/CVPR.2018.00221

Qin, R. (2014). *Change detection on LOD 2 building models with very high resolution spaceborne stereo imagery*. https://doi.org/10.1016/J.ISPRSJPRS.2014.07.007

Qin, R. (2019). A Critical Analysis of Satellite Stereo Pairs for Digital Surface Model Generation and a Matching Quality Prediction Model. *ISPRS Journal of Photogrammetry and Remote Sensing*, *154*, 139–150. https://doi.org/10.1016/j.isprsjprs.2019.06.005

Scharstein, D., Szeliski, R., & Zabih, R. (2001). A Taxonomy and Evaluation of Dense Two-Frame Stereo Correspondence Algorithm. *Proceedings IEEE Workshop on Stereo and Multi-Baseline Vision (SMBV 2001)*, *47*, 131–140. https://doi.org/10.1109/SMBV.2001.988771

Song, X., Yang, G., Zhu, X., Zhou, H., Wang, Z., & Shi, J. (2021). AdaStereo: A Simple and Efficient Approach for Adaptive Stereo Matching. *2021 IEEE/CVF Conference on Computer Vision and Pattern Recognition (CVPR)*, 10323–10332. https://doi.org/10.1109/CVPR46437.2021.01019

Yuan, W., Zhang, Y., Wu, B., Zhu, S., Tan, P., Wang, M. Y., & Chen, Q. (2021). Stereo Matching by Self-supervision of Multiscopic Vision. *ArXiv:2104.04170 [Cs]*. http://arxiv.org/abs/2104.04170

Žbontar, J., & LeCun, Y. (2015). Computing the Stereo Matching Cost with a Convolutional Neural Network. *2015 IEEE Conference on Computer Vision and Pattern Recognition (CVPR)*, 1592–1599. https://doi.org/10.1109/CVPR.2015.7298767

Zhong, Y., Dai, Y., & li, H. (2017). *Self-Supervised Learning for Stereo Matching with Self-Improving Ability*.

Zhou, C., Zhang, H., Shen, X., & Jia, J. (2017). *Unsupervised Learning of Stereo Matching*. 1567–1575. https://openaccess.thecvf.com/content_iccv_2017/html/Zhou_Unsupervised_Learning_of_ICCV_2017_paper.html